\newcommand{\CLASSINPUTtoptextmargin}{19mm}
\newcommand{\CLASSINPUTbottomtextmargin}{18mm}

\documentclass[letterpaper, 10 pt, conference]{ieeeconf}

\IEEEoverridecommandlockouts
\usepackage{cite}
\usepackage{amsmath,amssymb,amsfonts}
\usepackage{graphicx}
\usepackage{textcomp}
\usepackage{xcolor}

\usepackage{algorithm}
\usepackage[noend]{algpseudocode}
\usepackage{subcaption}

\def\BibTeX{{\rm B\kern-.05em{\sc i\kern-.025em b}\kern-.08em
    T\kern-.1667em\lower.7ex\hbox{E}\kern-.125emX}}

\title{\LARGE \bf
Selective Self-Assembly using Re-Programmable Magnetic Pixels
}

\author{Martin Nisser$^{1}$
Yashaswini Makaram
\thanks{$^{1}$Authors are with MIT CSAIL
        {\tt\small nisser@mit.edu}}%
}

\author{Martin Nisser$^{1}$ \hspace{0.3cm} Yashaswini Makaram$^{1}$ \hspace{0.3cm} Faraz Faruqi$^{1}$ \hspace{0.3cm} Ryo Suzuki$^{2}$ \hspace{0.3cm} Stefanie Mueller$^{1}$
\thanks{$^{1}$Authors are with MIT CSAIL
        {\tt\small \{nisser@mit.edu\}}}%
\thanks{$^{2}$Author is with University of Calgary}
}

\begin{document}





\maketitle

\begin{abstract}
This paper introduces a method to generate highly selective encodings that can be magnetically "programmed" onto physical modules to enable them to self-assemble in chosen configurations. We generate these encodings based on Hadamard matrices, and show how to design the faces of modules to be maximally attractive to their intended mate, while remaining maximally agnostic to other faces. We derive guarantees on these bounds, and verify their attraction and agnosticism experimentally. Using cubic modules whose faces have been covered in soft magnetic material, we show how inexpensive, passive modules with planar faces can be used to selectively self-assemble into target shapes without geometric guides. We show that these modules can be easily re-programmed for new target shapes using a CNC-based magnetic plotter, and demonstrate self-assembly of 8 cubes in a water tank.

\end{abstract}


\section{Introduction}



Modular self-assembly is a parallel process for the bottom-up fabrication of arbitrarily complex geometries from a set of disjoint modules. At the molecular scale, self-assembly is effectively used in nature to assemble complex biological structures \cite{whitesides1991molecular}, and formal models evaluating the power of self-assembly have proven it to be Turing-universal. Seeking to harness its capabilities, engineers have used self-assembly across scales to create microstructured materials with designed optical and mechanical properties at the $\mu$m-scale \cite{grzybowski2003electrostatic}, to self-assembled structures at the mesoscale~\cite{nisser2016feedback} up to aerial self-assembly of structures at the m-scale \cite{papadopoulou2017self}. As an alternative to automating the assembly of robots in a top-down manner~\cite{nisser2021laserfactory},  reconfigurable robotics has explored bottom-up self-assembly in two forms, active and passive. 


Active self-assembly by modular self-reconfigurable robots (MSRR) involves modules that modulate their behavior online in order to locate, position and bond themselves to their neighbors, for which each module requires embedding with computation, sensing and actuation. Because most systems choose cubic modules to facilitate tessellation, a key discriminator between systems is often the actuation, which include electropermanent magnets \cite{hauser2020kubits}, electromagnets \cite{nisser2017electromagnetically,nisser2022electrovoxel}, momentum wheels \cite{romanishin2013m, romanishin20153d}, and COTs servos or motors \cite{baca2014modred}. While active assembly has been used to successfully reconfigure a variety of robotic systems, it is these actuators that are typically the most significant challenge to scaling systems up in number and down in size due to the cost and complexity of embedding them into individual modules \cite{zykov2007experiment}.

\begin{figure}
  \centering
  \includegraphics[width=1\columnwidth]{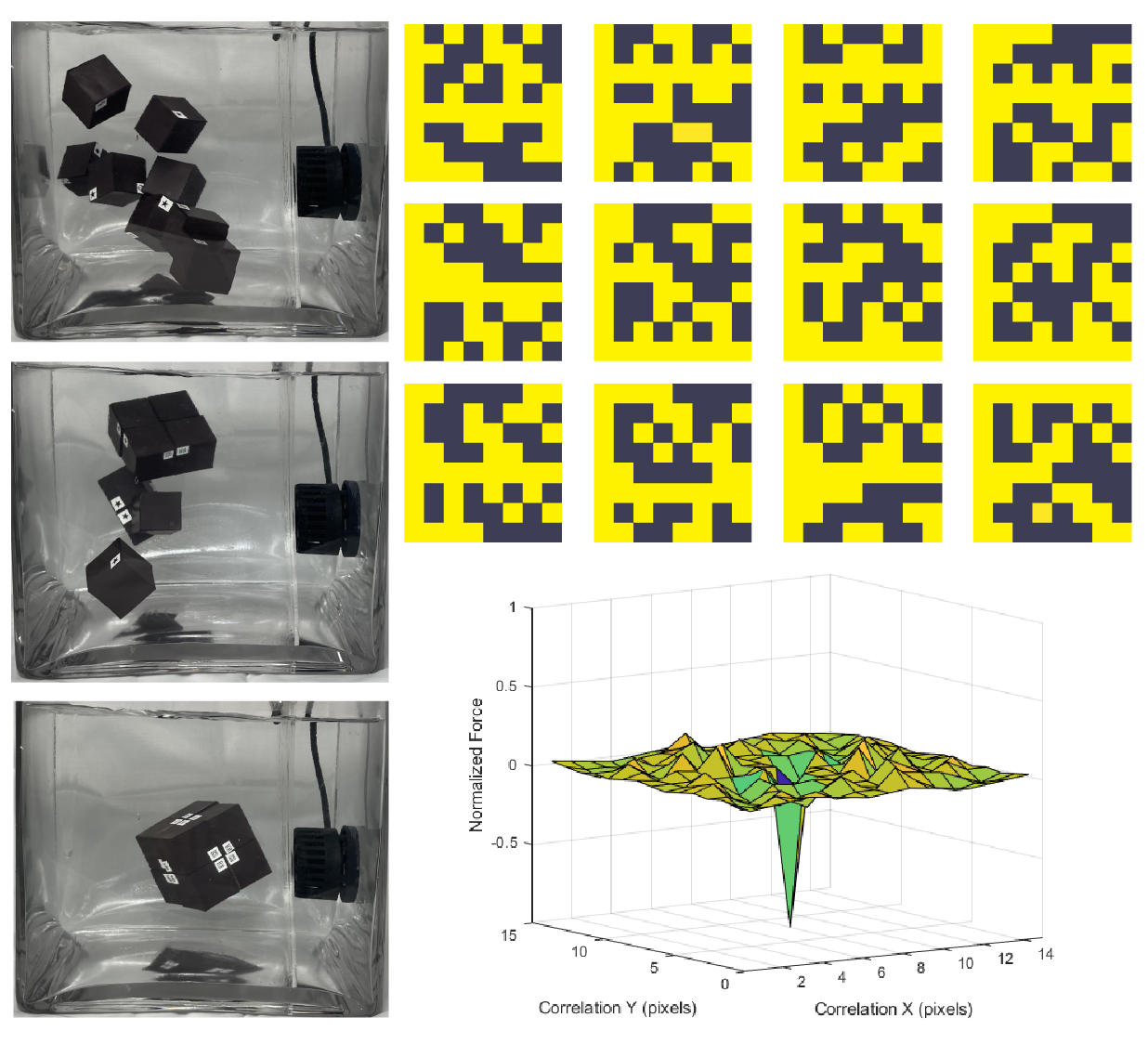}

\caption{(Left) We enable stochastic self assembly using inexpensive (\$0.23) cubic modules. (Above, right) We accomplish this by magnetically programming module faces with uniquely mating pairs of encodings based on Hadamard matrices, and show bounds on their performance. (Below, right) Key to the modules' success is their ability to attract strongly to their mates, while remaining agnostic in all other translations, rotations, and to non-mating modules.}
  \label{fig:teaser}
\vspace{-0.6cm}
\end{figure}


\begin{figure*}
  \centering
  \includegraphics[width=1\textwidth]{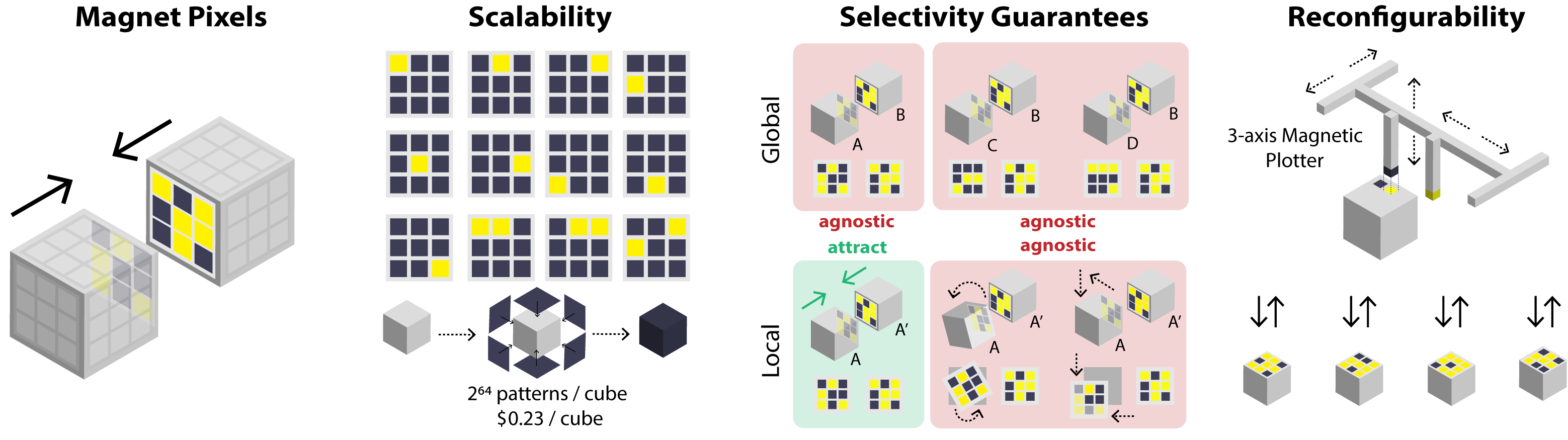}
\caption{Overview of our stochastic self assembly approach. (Left) \textit{Magnetic Pixels:} Cubic modules are programmed with matrices of magnetic pixels. These permit module faces to mate selectively to assemble target geometries. 3x3 Matrices shown for clarity; our modules utilize 8x8. (Center left) \textit{Scalability:} Our binary-valued, 8x8 matrices can encode 2$^{64}$ module faces with unique permutations, and modules are inexpensive (\$0.23). (Center right) \textit{Selectivity Guarantees:} We leverage Hadamard matrices to encode magnetic pixels on faces with 2 criteria. "Locally", mating faces attract in only one configuration; "globally", non-mating faces attract in no configuration. (Right) \textit{Reconfigurability:} Encodings are "programmed" as magnetic pixels using magnets installed on a 3-axis CNC. Modules can be re-programmed to self assemble into new target geometries.}
  \label{fig:overview}
\vspace{-0.6cm}
\end{figure*}

In contrast, passive self-assembly obviates the need for emdedded acuation and control. Instead, system actuation is outsourced to an external excitation, and in the case of \textit{stochastic} self-assembly~\cite{tibbits2012self}, this excitation requires no local control and is governed instead only by global parameters such as excitation magnitude. Stochastic assembly sacrifices efficiency and predictability for advantages in cost, complexity and scale; by enabling the environment to actuate reconfiguration, it trades off deterministic assembly times of individual modules for statistical assembly rates of the collective. To assemble stochastically, modules require pre-programming to enforce correct mating during random collisions with their intended mate. This programmed specificity between pairs of mating faces is typically achieved via minimization of free surface energy via topology\cite{hacohen2015meshing}, wettability \cite{bowden1997self}, magnetic forces \cite{lu2021enumeration} or electrostatic \cite{grzybowski2003electrostatic} interaction. While the wider scientific community has often been interested in constraining the self-assembly problem to 2D, for instance by using a shaker table \cite{jilek2020centimeter}, roboticists have leveraged liquid tanks to study assembly in 3D. Fluidic assembly at the mesoscale has become a particularly widely studied problem in robotics \cite{tolley2008dynamically,tolley2010fluidic,tolley2011programmable,krishnan2008increased,kalontarov2010hydrodynamically, zykov2007experiment}. Existing stochastically self-assembling modules typically include two features to enable assembly: first, embedded magnets that generate near-field forces to bring modules close, and second, selective geometry on module faces that encodes the specificity to only permit bonds between mating pairs \cite{jilek2021towards,hacohen2015meshing,jilek2020centimeter, tsutsumi2007multistate}. However, three key challenges remain for the development of stochastic self-assembling systems: (1) \textit{scalability} that shows how modules can be made both numerous and small; (2) \textit{selectivity guarantees} that help bound module misassembly; and (3) \textit{reconfigurability} that let modules acquire different target shapes. 

\textit{(1) Scalability:} To assemble arbitrarily complex geometries, encodings for 3D modules must support selectivity great enough to permit uniquely mating pairs of modules in the hundreds or even thousands. In addition, modules must be inexpensive and simple enough to be fabricated in these quantities. Due to this dual problem, a significant corpus of previous research demonstrates the stochastic self-assembly for tiled 2D arrays, such as chessboards\cite{grzybowski2003electrostatic,jilek2021towards,miyashita2009influence}, with only two module types where each module is selective to entirely half of all modules in the set. On the other hand, the individual fabrication of heterogeneous module topologies with manually embedded permanent magnets poses a significant challenge to scalable fabrication. 

\textit{(2) Selectivity guarantees:} Because magnet arrangements typically used to generate near-field forces are poorly discriminating to each other, this framework often leads to misassemblies, because near field forces between both mating and non-mating face magnets are equally strong. In addition, protruding geometrical features used for selectivity can lead to "jamming" by obstructing assembly paths \cite{jilek2021towards}, and bounds on the expected misassembly rate between geometrically dissimilar modules may be difficult to compute.


\textit{(3) Reconfigurability:} To date, structures self-assembled at the mesoscale are not reconfigurable. Because module selectivity is achieved by fabricating individualized geometries, any set of fabricated modules encode only a single target shape (or single set, for non-deterministic encodings). Such modules are therefore unable to be "re-programmed" to self-assemble new target shapes: new shapes require a unique batch of modules to be fabricated from scratch, inhibiting their utility and increasing their potential unit cost.

In this paper, we introduce a method to design and "program" selective encodings~\cite{nisser2022stochastic} onto cubic modules in a way that addresses all three challenges above (Fig. \ref{fig:overview}). We program module faces with patterns of magnetic pixels which can attract or repulse pixels of another face (Fig. \ref{fig:overview} left), and if the number of pixels in attraction match those in repulsion, the faces are agnostic to each other. Using this observation, we show how to program modules with encodings that allow them to selectively mate with other cubes to self assemble in a unique target structure. Formulated as matrices, we demonstrate the number of unique encodings that can be programmed given criteria on attraction and agnosticism. 

Our modules consist of PLA cubes, with squares of soft magnetic material (inexpensive COTS fridge magnets) bonded to their sides (Fig. \ref{fig:overview} center left). The encodings on these faces generate both near-field forces \textit{and} selectivity. This selectivity lets us use homogeneous cubic modules with planar faces, making modules both non-jamming and inexpensive to mass fabricate across scales. Key to our approach is the generation of selective encodings, for which we leverage Hadamard matrices (Fig. \ref{fig:overview} center right), and similar procedures may have been used to create industrial-grade magnets with tailored selectivity properties in industry~\cite{CorrMag}. The two polarities of the magnetic pixels we encode onto faces correspond to elements of these binary-valued matrices. Our matrices enforce two key criteria. A local criterion dictates behavior between faces intended to mate. If every pixel on one face, face A, is magnetically opposite to pixels on another, then we call that face its mate, A'. These faces thus form a maximally attractive mating pair. Our matrix pairs are designed to exhibit maximal attraction in this one configuration, while placing guarantees on agnosticism in all other translations and rotations. A global criterion dictates behavior between two faces not intended to mate. For this, we ensure that any given matrix pair, A and A', cannot mate with any other faces B, B', C, C' etc., in any configuration, and we place guarantees on this agnosticism. We further demonstrate how to "program" module faces using a simple magnetic plotter consisting of two oppositely oriented permanent magnets affixed to a 3-axis CNC (Fig. \ref{fig:overview} right). Crucially, these soft magnetic faces are re-programmable, and thus modules can be repeatedly re-programmed with new encodings in order to self-assemble into new target geometries in 3 dimensions. Finally, we design and fabricate a set of 8 modules and demonstrate their stochastic self-assembly in a water tank.

The paper is structured as follows. We begin by introducing a procedure to generate selective magnetic encodings, and derive bounds on the number of modules that can be utilized given a threshold on agnosticism between programmed encodings. We describe the physical modules themselves, and the magnetic plotting technique used to program them. We demonstrate our ability to make predictions with regard to the attraction and agnosticism between various magnetic encodings, and verify these experimentally. Finally, we demonstrate stochastic self-assembly of our system using 8 modules in a water tank.

\section{Encoding generation}


This section describes how we generate encodings that satisfy the global and local criteria given above. Our encodings are based on Hadamard matrices, whose unique properties have lent their use to applications including Code Division Multiple Access and error correcting code. The Hadamard matrix is a square matrix whose rows are all mutually orthogonal and whose elements are either $1$ or $-1$ (here representing magnetic pixel polarization). As a consequence of its row orthogonality, it follows that its columns are mutually orthogonal too. As a result, the dot product of any pair of rows, or any pair of columns, is equal to 0. In addition, if a single row or column is multiplied by -1 before taking the dot product, the product remains 0. 

Defining the mate of matrix A as A'$=$-A, the Hadamard product (\ref{eq_hada_prod}) between a matrix A of order N and its mate, normalized by the number of elements N$^2$, is -1. This implies maximum magnetic attraction. Conversely, the normalized Hadamard product between A and itself is +1, connoting maximal repulsion. Let $S_G$ be a global score indicating the maximum attraction between non-mating pairs A and B in any configuration, and let $S_L$ be the local score indicating maximum attraction between mating pairs A and A' \textit{in all wrong configurations}. To permit self assembly, satisfying the global and local agnosticism criteria requires the force enacted by our fluid $F_f$ to satisfy $-1 < F_f < min(S_L,S_G)$ in order to both break apart unintended misassemblies and allow correct assemblies to survive. Because controlling the fluidic force is challenging, the goal is then to maximize $min(S_L,S_G)$, making them maximally agnostic, in order to place $F_f$ between these values. 

Now, the row and column orthogonality described above yields that taking the Hadamard product between A and A' becomes 0 if one matrix is translated in only X or only Y, yielding maximal agnosticism. However, agnosticism is not guaranteed for matrices translated in both X \textit{and} Y, or if they are rotated. To find matrices that maximize agnosticism in these configurations, we perform two searches. The first search shows that Hadamard matrices perform optimally among the set of all square matrices. The second search identifies the number of matrices that can be generated for a given bound on agnosticism performance.

\begin{equation}
   (A \odot B)_{ij} = (A)_{ij} (B)_{ij}
\label{eq_hada_prod}
\end{equation}

\subsection{Matrix search}

We generate every binary-valued (values 1 and -1) matrix A of order N=4 (yielding $2^{16}$ matrices), and compute the cross-correlation between A and A', which is equivalent to their Hadamard product in every translation ($2N-1$ values) and also compute their Hadamard product in rotations of 0$^{\circ}$, 90$^{\circ}$, 180$^{\circ}$, and 270$^{\circ}$ (4 values, upper-bounded by computational resources). These values are used to assess the local criterion. We then check which of these are Hadamard matrices by evaluating which matrices A satisfy $AA^T = N(I_N)$, finding 768 matrices (1.2\%) out of the total 65,536 ($2^{16}$). We then rank every matrix by its most attractive (most negative) value for all translations and rotations (excluding its one mating configuration of -1) to give its local criterion score $S_L$. An optimal $S_L$ score is 0, implying perfect agnosticism in all configurations besides its mating configuration.

Assessing a set of matrices' global criterion score, $S_G$, requires finding the worst (most negative) Hadamard product in translation and rotation of every combination of matrices in the set. This is the power set of matrices, $2^{65,536}$ combinations for the matrices above, a computationally intractable search; searching for all possible cliques (maximal complete subgraphs) in a graph is an NP-complete problem. However, any guarantee of a matrix' performance in agnosticism is given by $min(S_L,S_G)$. Therefore, to narrow our search, we build sets by searching those with best $S_L$ first, stopping our search once $S_G=S_L$. We consider a graph $G$ where vertices represent these matrices, and where two vertices ($V_1, V_2$) are adjacent if their Hadamard product (in translation and rotation) is above some $S_G$. Hence we reduce the problem of searching for mutually compatible sets in the power set to a search for cliques in G. We determine $S_G$ for progressively larger clique sizes of the 65,536 matrices using a variation of the algorithm by Bron and Kerbosch~\cite{bron1973algorithm,hagberg2008exploring}, and find that given a $S_G$ and $S_L$, the set of Hadamard matrices produce a clique size equivalent to the set of all possible matrices. Thus, the maximal clique search, which exhibits exponential complexity, is reduced to the set of Hadamard matrices (a small fraction, 1.2\%, of the total matrices possible) and still finds maximally sized cliques given some $S_G$ and $S_L$. 


Initial experiments revealed that matrices of order N=4 were unable to produce adequately agnostic values for $S_G$ and $S_L$ for clique sizes greater than 5 such that our stochastic fluid could discriminate between correct and incorrect mates. However, with a goal to assemble an 8-module system into a "meta cube" (Fig. \ref{fig:assembly}), we require a minimum clique size of 12. We therefore use our result regarding maximal cliques to search across Hadamard matrices of order N=8.



\subsection{Generating Hadamard Matrices}

We generate a normalized Hadamard matrix of order N=8 using the recursive procedure described below: let $H$ be a Hadamard matrix of order N. We can use this to create the partitioned matrix of order $2N$ shown in (\ref{eq1}).

\begin{figure}
  \centering
  \includegraphics[width=0.98\columnwidth]{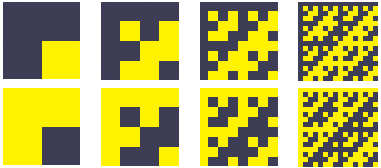}
\caption{(Above) Normalized, naturally ordered Hadamard matrices generated in a procession of orders N of 2, 4, 8 and 16. Binary values of 1 and -1 are represented as dark and light pixels, respectively. (Below) Their mates.}
  \label{fig:hadamard}
\end{figure}

\begin{equation}
\begin{bmatrix}
H & H \\ H & -H
\end{bmatrix}
\label{eq1}
\end{equation}
We can apply this rule generally for higher orders using:
 \begin{equation}
H_{2^k} = 
\begin{bmatrix}
H_{2^{k-1}}  &  H_{2^{k-1}} \\ 
H_{2^{k-1}}  & -H_{2^{k-1}}
\end{bmatrix}
\label{eq2}
\end{equation}
Initializing $H_1$ as 1, we generate $H_8$ and see that the Hadamards produced in this construction are fractal (Fig. \ref{fig:hadamard}). We then generate all $8!$ matrices that are permutations of its rows. As the rows of $H_8$ are orthogonal, the rows of all its permutations remain orthogonal and Hadamard.

 

\begin{figure}
  \centering
  \includegraphics[width=0.98\columnwidth]{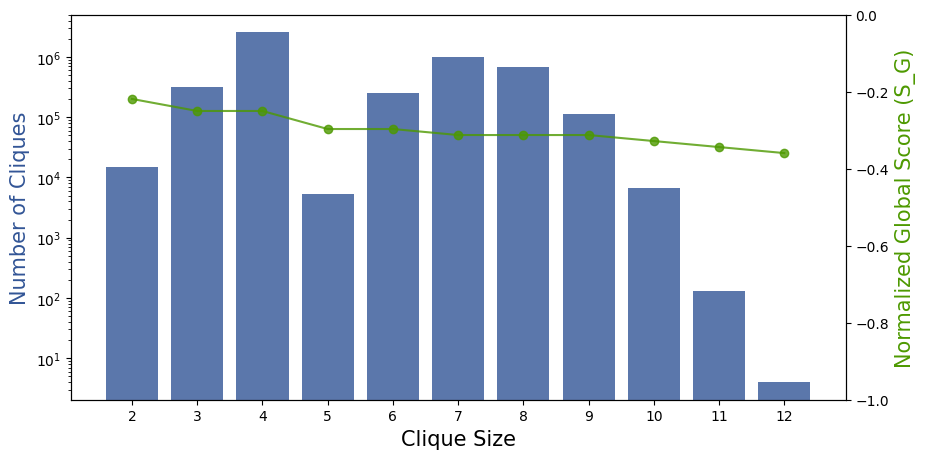}
  \caption{The size of a clique (the number of mutually compatible encodings) related to both 1) the number of such cliques, and 2) its global agnosticism score $S_G$}
  \label{fig:clique_size_score}
\end{figure}

To meet our goal of finding a clique of size 12, we seed a threshhold for both $S_L$ and $S_G$ of -0.2, and iteratively search for maximal cliques. We continue to lower the threshold of $S_G$ by 0.02 until we grow the maximal clique size to 12. Fig. \ref{fig:clique_size_score} illustrates the relationship between maximum clique size, its mutual $S_G$ score, and the number of cliques of that size. We select the 12 matrices from one of the 4 size-12 cliques to program and assemble a meta cube, exhibiting a combined agnosticism score of $min(S_G,S_L)=S_G=-0.36$.

\section{Magnetically programmed modules}

\begin{figure}
  \centering
  \includegraphics[width=0.9\columnwidth]{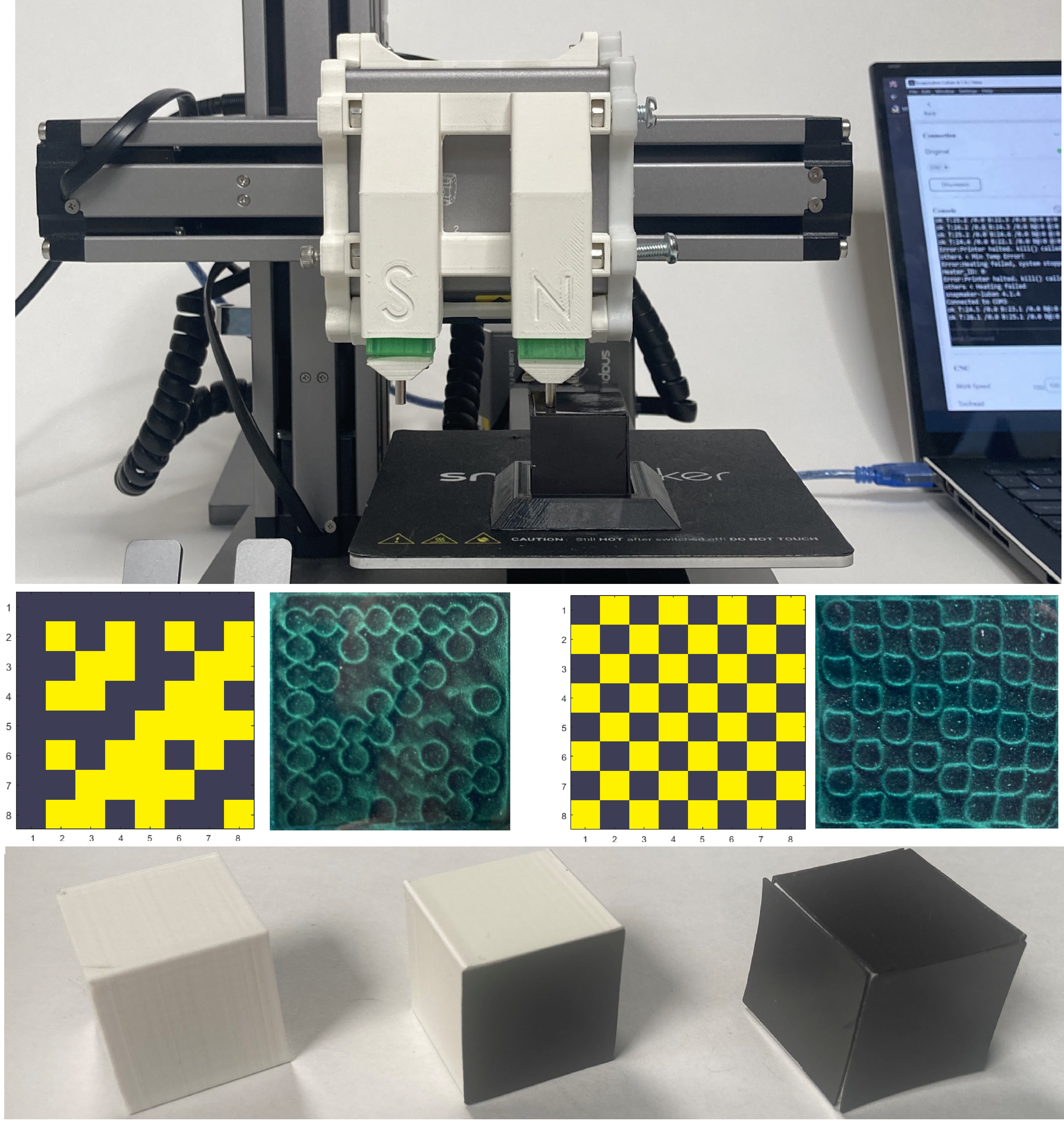}
  \caption{(Above) Magnetic plotter programming a module. (Middle) Simulated and plotted matrices viewed through magnetic viewing film; (Left) A normalized order-8 Hadamard; (Right) A checkerboard. Dark and light pixels in simulation represent opposite magnetic polarities. The same material was reprogrammed to produce these patterns in turn. (Below) Module fabrication. (Left) a white PLA cube is covered with (middle) squares of black soft magnetic material; (Right) once all six squares are bonded, it is ready for programming.}
  \label{fig:plotter}
\end{figure}

Soft magnetic material is material that can be easily coerced to become magnetized when placed in an external magnetic field. When removed from the field, they nonetheless retain a significant fraction of their magnetization, letting them function as magnets. This process is repeatable, allowing modules with soft magnetic faces to be re-programmed with new encodings that encode different target structures. 

We build a magnetic plotter to stamp our cubes' soft magnetic faces with magnetic encodings (Fig. \ref{fig:plotter}, above). The plotter consists of a pair of oppositely polarized permanent magnets (3mm diameter, 6mm length) installed in a housing mounted onto a 3-axis CNC (SnapMaker 3-in-1). Each magnet thereby plots opposite pixel values, where binary-valued pixels correspond to oppositely polarized regions\textemdash magnetic pixels\textemdash of soft magnetic material. A script translates these matrices into G-code, allowing the plotter to program faces without manual intervention. Once plotted, encodings can be viewed using magnetic viewing film (Fig. \ref{fig:plotter}, middle).

We 3D print 25mm cubes from PLA and bond square faces of 26-mil thick soft magnetic material to its 6 faces (Fig. \ref{fig:plotter}, below). The cubes are printed with internal cubic cavities of side length 18mm to neutrally buoy them in tap water. Cubes are placed in the CNC platform and programmed with encodings that produce the desired target configuration once mated. Each module costs \$0.23 in materials (\$0.19 PLA, \$0.04 for 6 soft magnetic squares), and requires 12 minutes to program all 6 faces with our un-optimized G-code. Taken together, cubes programmed with these encodings are therefore inexpensive, easy to manufacture, physically re-programmable, attractive over distances in contrast to contact adhesives (such as glue), and generate forces selectively without consuming power during operation.

\section{Results}

In this section, we measure the magnetic force of the programmed encodings in attraction and repulsion, and compare this to predictions made by taking Hadamard products of the associated matrices. We then evaluate the performance of the matrices in terms of the global and local agnosticism criteria.

\begin{figure}[h]
  \centering
  \includegraphics[width=0.85\columnwidth]{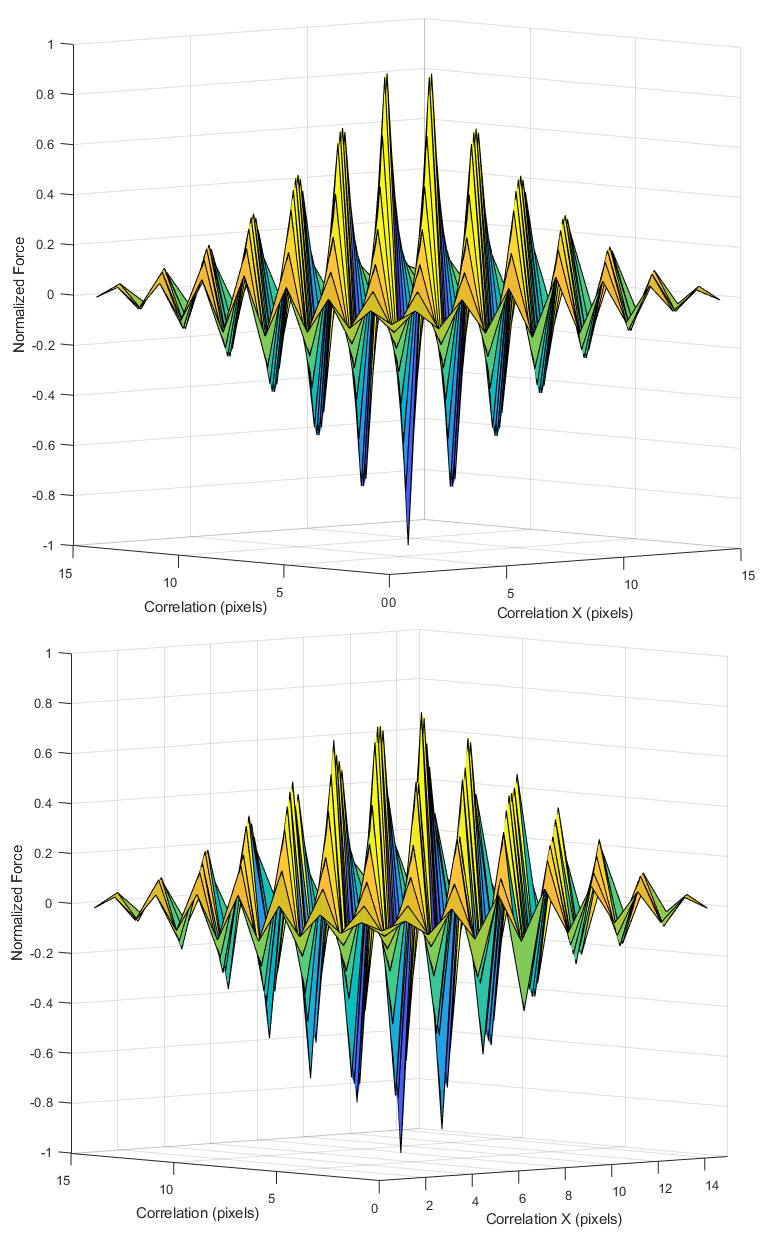}
  \caption{(Above) Predicted vs (below) measured magnetic force for a checkerboard matrix translated with its mate.}
  \label{fig:theory_meas}
\end{figure}

\subsection{Empirical validation}

We place one programmed cube on a scale (KUBEI pocket, 0.1mN accuracy) installed on the CNC platform, and mount a second cube above it on the CNC end effector, aligning the mating faces to be planar at 0.5mm distance in Z. We use the CNC to translate the second cube one pixel at a time, in X and Y, and measure the force generated at each increment. To maximise signal to noise, the cubes are programmed with mating checkerboard patterns of size N=8 (Fig. \ref{fig:plotter}, middle right). This is done because unlike the encoding matrices which are largely agnostic, checkerboard patterns generate large cyclic forces in attraction and repulsion as like and unlike pixels align and misalign with each translation. The predicted results are generated by taking the cross-correlation between mating checkerboard matrices, i.e. the matrices are translated pixel-wise while taking their Hadamard product (\ref{eq_hada_prod}) at each increment. The correlations in this and subsequent figures are plotted normalized by the peak attractive (negative) force in order to facilitate comparison between predicted and measured results.

Fig. \ref{fig:theory_meas} shows the predicted (above) and measured (below) results, which match well visually. However the measured repulsive values are weaker than those predicted by correlating the matrix values. We calibrated the scale to rule out ascribing this result to anisotropic measurement sensitivity. Rather, this is likely an effect of coercivity of the pixels on each other; if two attractive magnetic dipoles are brought into contact, they reinforce their attractive alignments. However, repulsive dipoles will realign to an attractive equilibrium if free to rotate; a condition which the low coercivity of the soft magnetic faces may support. Accounting for this, we implement a scaling factor of 0.09 to the repulsive forces predicted by correlation that normalizes the magnitude of the repulsive pixels to those in attraction. We compute a normalized sum of squared differences of 0.014 between the measured and predicted results using this scaling factor, supporting our model as an accurate predictor of force between magnetically programmed faces. 

\subsection{Local agnosticism criterion}
In this section, we use our model to evaluate the success of the local agnosticism criterion; the forces between our generated matrices and their mates. The figures are emblematic of the performance of all matrices in our clique.

\begin{figure}[h]
  \centering
  \includegraphics[width=0.92\columnwidth]{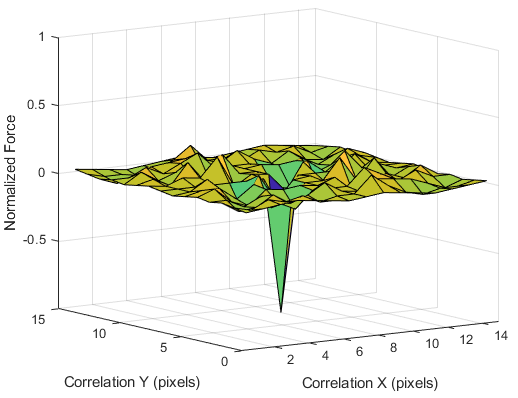}
  \caption{Translational agnosticism of a matrix A with its mate A'. They exhibit maximal attraction (-1) when centered, remaining largely agnostic (0) elsewhere.}
  \label{fig:correlation-local-trans}
\end{figure}

To evaluate selectivity in translation, we correlate the matrices in X and Y, taking the Hadamard product at each pixel increment (Fig. \ref{fig:correlation-local-trans}). A peak normalized attractive force of -1 is produced with the matrices translationally centered at (0,0); this corresponds to 256 Pascals, or 160 mN between square faces of side 25mm. Elsewhere, the correlation is dominantly agnostic (centered about 0) or repulsive (positive), bounded in attraction by -0.25. 

\begin{figure}[h]
  \centering
  \includegraphics[width=0.95\columnwidth]{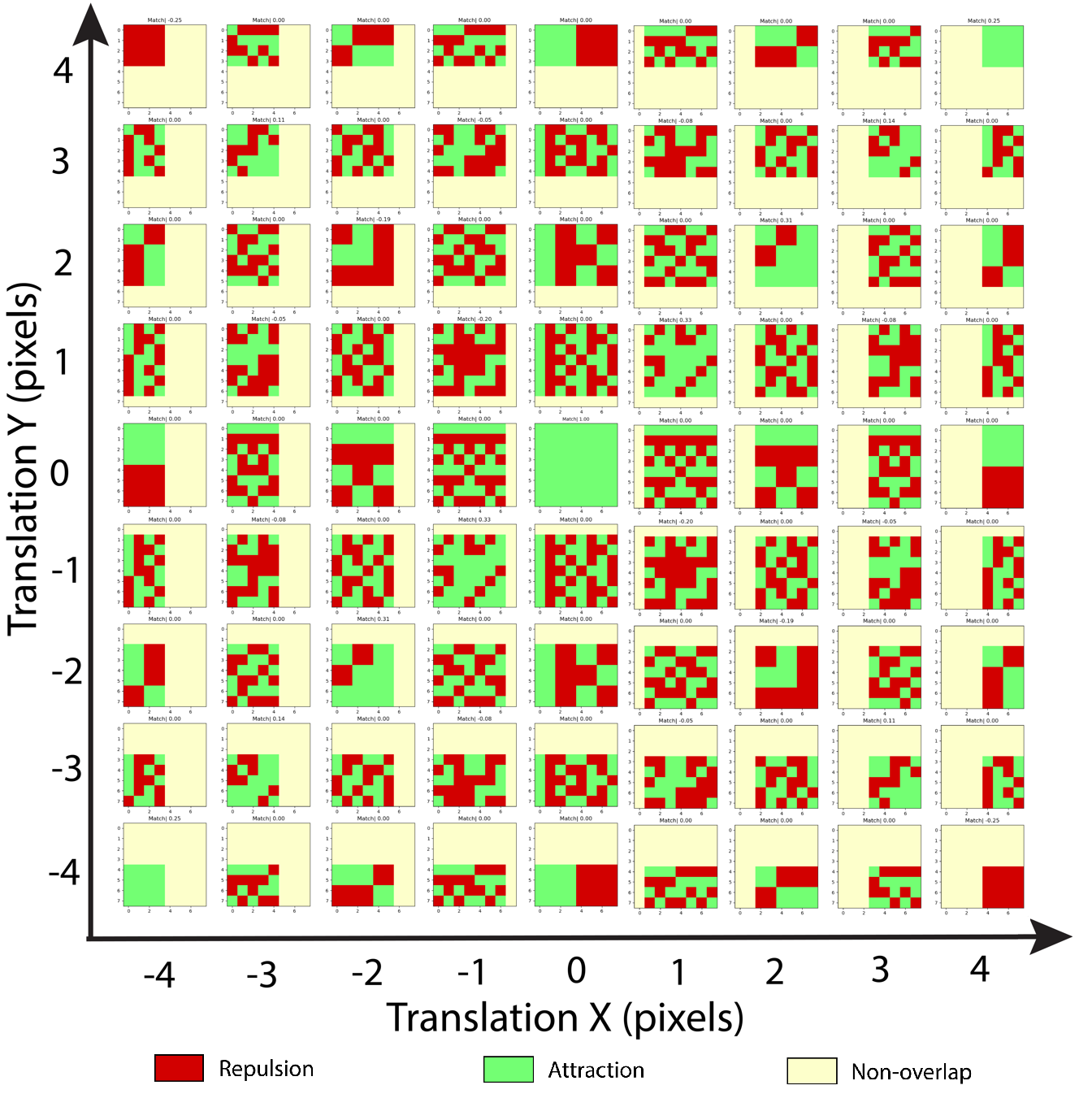}
   \caption{Pixel-wise attraction and repulsion of a normalized order-8 Hadamard matrix during translation with its mate. These are maximally attractive with the matrix pair translationally centered at (0,0), during full overlap. Elsewhere, attractive and repulsive forces largely cancel. 
   }
   \label{fig:translation-viz}
\end{figure}

To illustrate how the Hadamard product produces a dominantly agnostic interaction between the matrices besides their mating configuration, Fig. \ref{fig:translation-viz} visualizes the pixel-wise attraction and repulsion during the correlation of a normalized Hadamard with its inverse. Here, red pixels indicate repulsion (+1), green attraction (-1), and yellow non-overlap (0). Summing the pixels over a square produces the Hadamard product, or the net force, that is plotted in each data point in Fig. \ref{fig:correlation-local-trans}. The translationally centered position (0,0) in Fig. \ref{fig:translation-viz} produces uniform attraction, whereas other positions produce an exactly or largely agnostic interaction due to equal numbers of attractive or repulsive pixels cancelling out. 

\begin{equation}
\begin{bmatrix}
x'\\y'
\end{bmatrix} = 
\begin{bmatrix}
cos(\theta) & -sin(\theta)\\ sin(\theta) & cos(\theta)
\end{bmatrix}
\begin{bmatrix}
x\\y
\end{bmatrix}
\label{eq_rot}
\end{equation}

\begin{figure}[t]
  \centering
  \includegraphics[width=0.97\columnwidth]{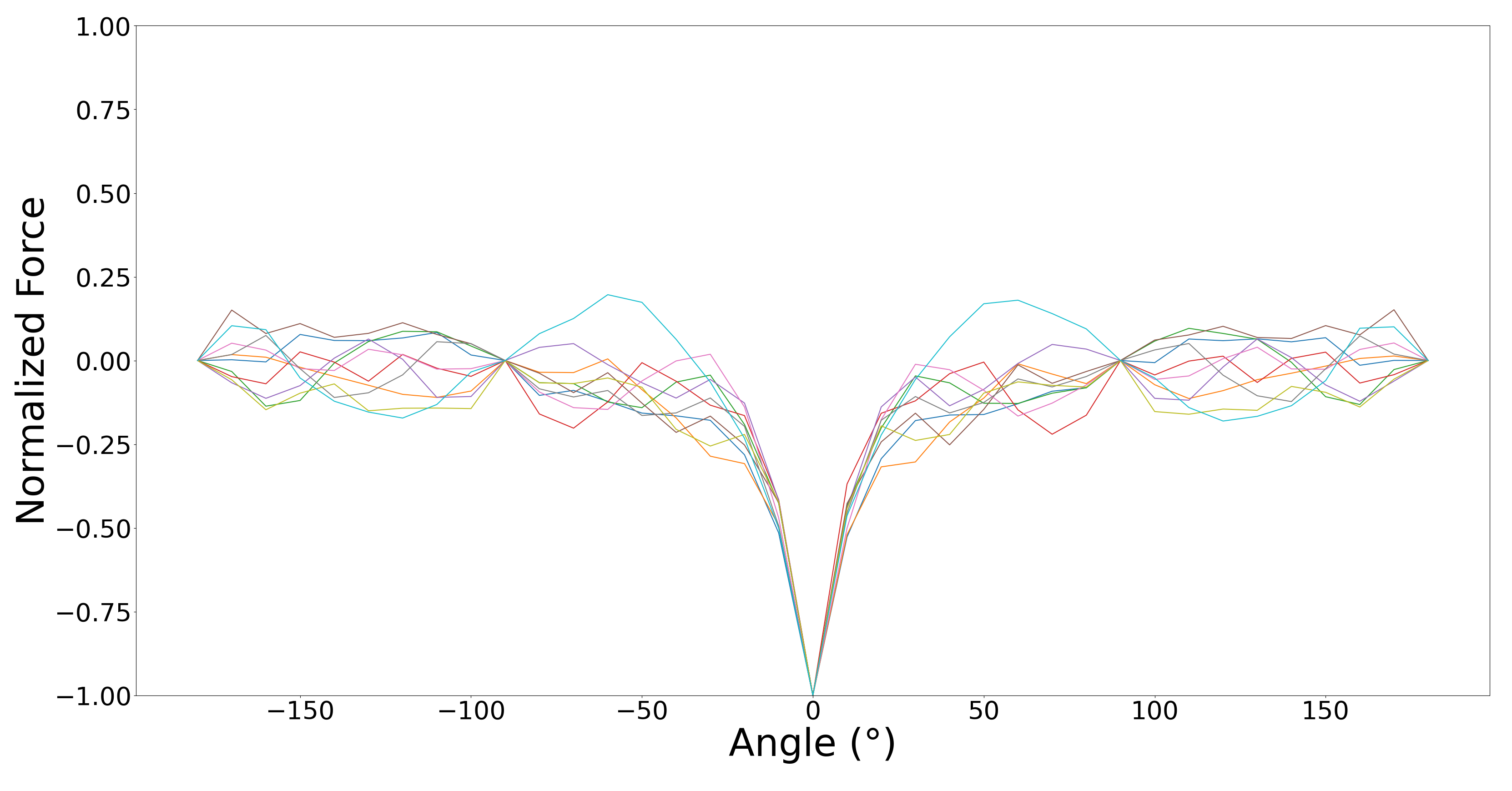}
  \caption{Rotational agnosticism of a matrix A with its mate A'. They exhibit maximal attraction (-1) when centered, remaining largely agnostic (0) elsewhere.}
  \label{fig:rotation-local}
\end{figure}

To evaluate selectivity in rotation, Hadamard products can be evaluated between the matrices for orientations of 90$^{\circ}n$, where n is an integer. Elsewhere, the matrices do not superimpose. Therefore to evaluate these products at arbitrary angles, we first discretize the order-8 matrices by a factor of 10 to produce 80x80 grids. Each matrix element is then said to be at location (x,y) with respect to an origin placed at the center of the matrix, and we use the rotation matrix (\ref{eq_rot}) to compute its new location (x',y') after an arbitrary rotation of $\theta$. We smooth the result using a 3x3 averaging kernel to remove artefacts from the discretization, then evaluate the Hadamard product to evaluate the net force, assigning a value of 0 (non-overlap, or agnosticism) where pixels do not overlap. Fig. \ref{fig:rotation-local} plots this force for 10 matrices from -180$^{\circ}$ to 180$^{\circ}$ in 10$^{\circ}$ increments. While our search procedure optimized only for rotations of multiples $90^{\circ}$, this procedure verifies performance for arbitrary rotations. The attractive force is bounded by -0.25 in all orientations besides the intended mating configuration (0$^{\circ}$). Pure agnosticism is further enforced at 90$^{\circ}n$, where the square matrices line up.

\begin{figure}[t]
  \centering
  \includegraphics[width=0.85\columnwidth]{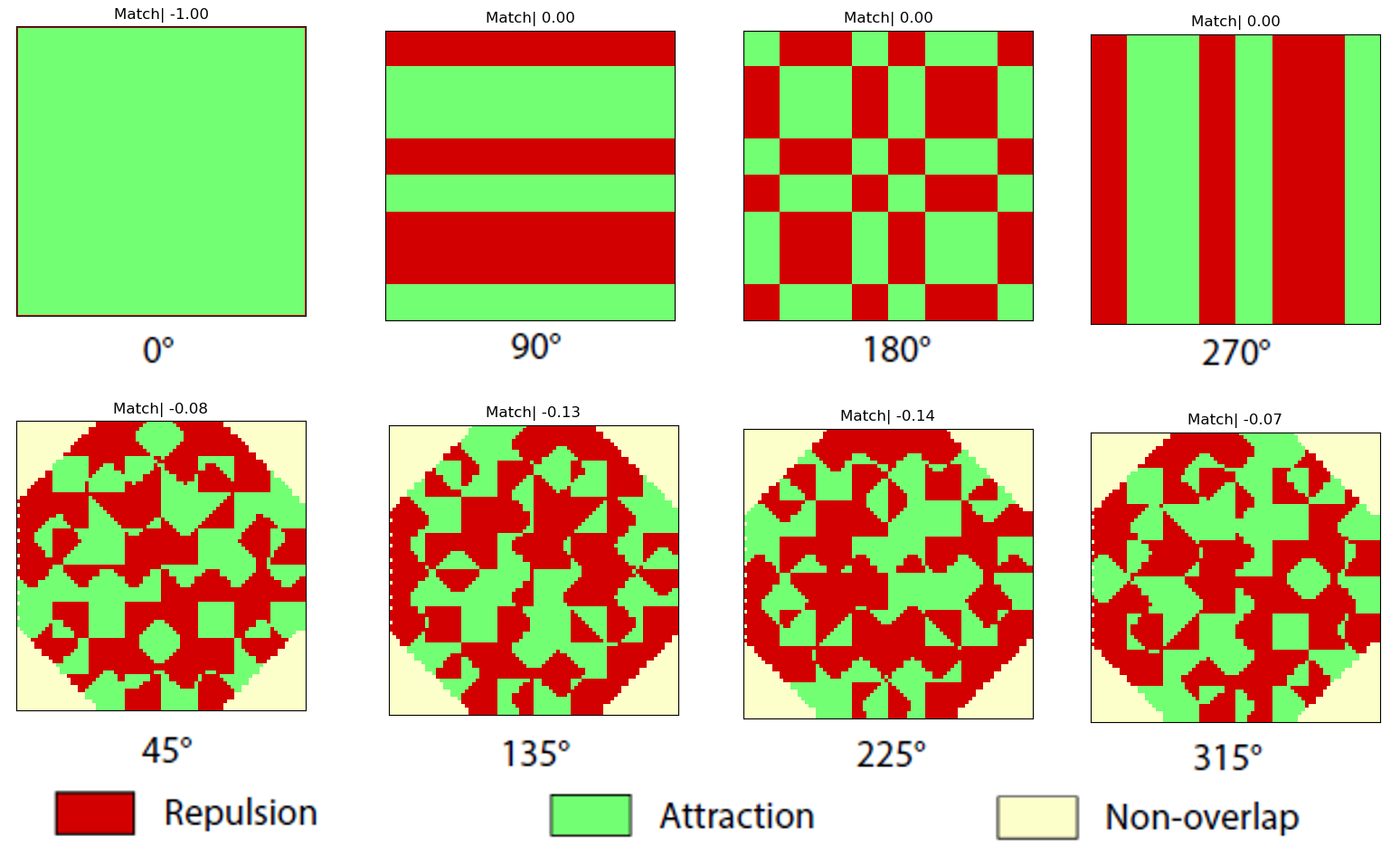}
   \caption{Pixel-wise attraction and repulsion of a normalized order-8 Hadamard matrix during rotation with its mate. For rotations of 90n$^{\circ}$, attraction and repulsion cancel exactly; elsewhere, approximately. }
   \label{fig:rotation-viz}
\end{figure}

Fig. \ref{fig:rotation-viz} visualizes the pixel-wise attraction and repulsion during the rotation of a normalized Hadamard with its inverse. At 0$^{\circ}$, 90$^{\circ}$, 180$^{\circ}$ and 270$^{\circ}$, locally attractive and repulsive pixels sum to 0, producing agnosticism. Elsewhere, attractive and repulsive pixel interactions cancel to within an attractive bound of -0.2.

\subsection{Global agnosticism criterion}

\begin{figure}[t]
  \centering
  \includegraphics[width=0.92\columnwidth]{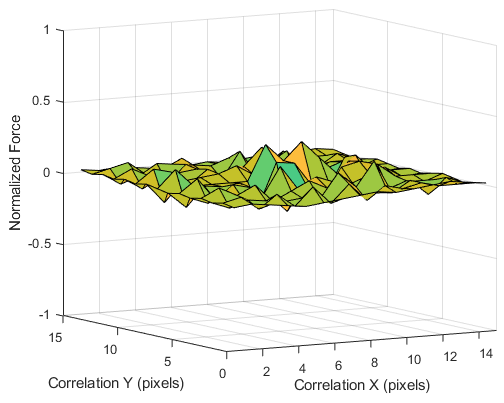}
  \caption{Translational agnosticism between non-mating matrices; agnosticism dominates for all configurations.}
  \label{fig:correlation-global}
\end{figure}

\begin{figure}[t]
  \centering
  \includegraphics[width=0.97\columnwidth]{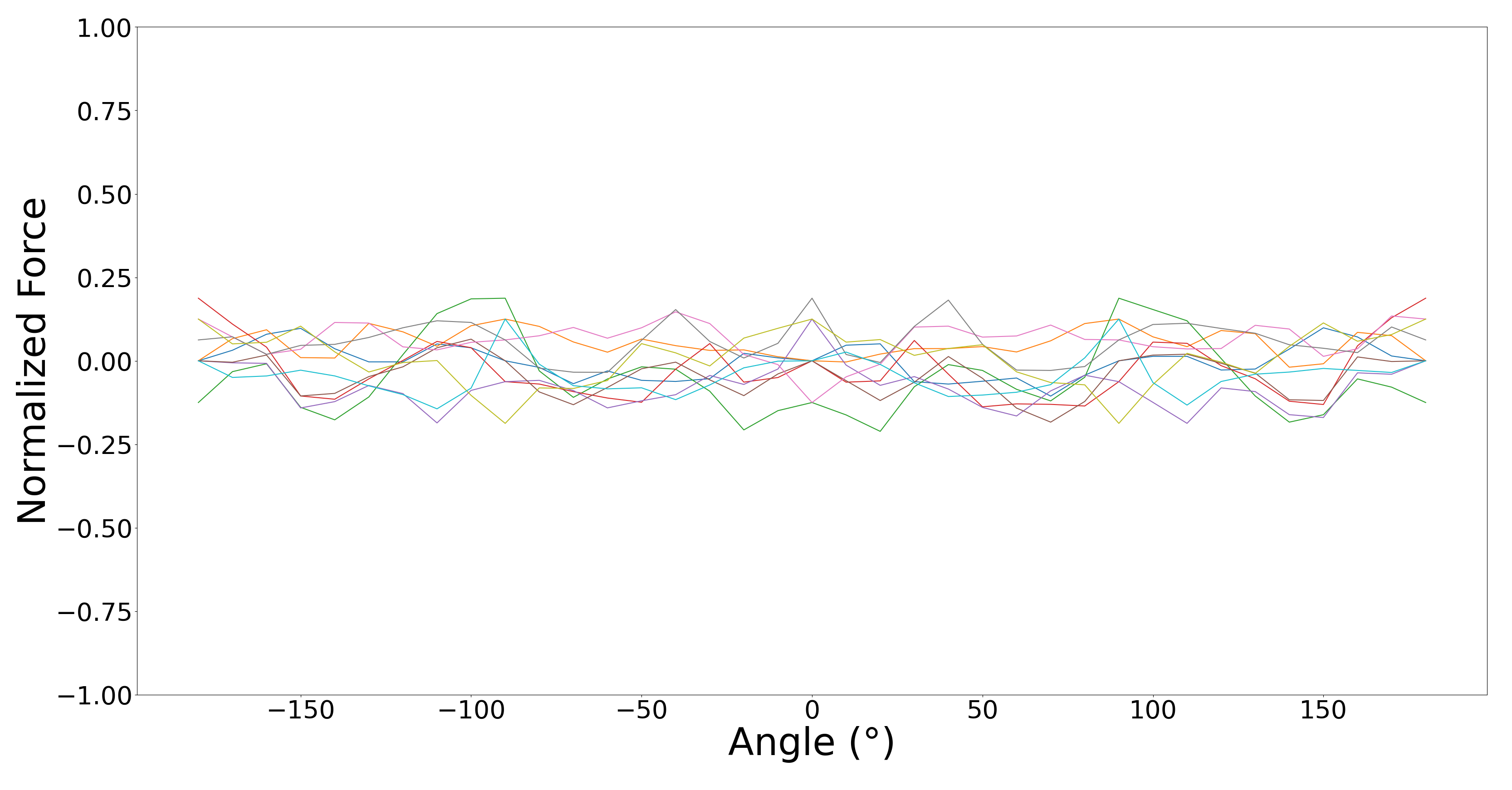}
  \caption{Rotational agnosticism between non-mating matrices; agnosticism dominates for all configurations.}
  \label{fig:rotation-global}
\end{figure}

In this section, we evaluate the global agnosticism criterion between two emblematic matrices from our clique. Using the same tools used in the local case above, Fig. \ref{fig:correlation-global} shows the correlation between two matrices, illustrating their agnosticism over all translations. Equivalently, Fig. \ref{fig:rotation-global} illustrates their agnosticism in rotation, with a negative bound of -0.36 that indicates that attraction between non-mating faces is never greater than 36\% of the attractive force between mating faces in alignment, as derived in our search.      

\subsection{Stochastic self-assembly}

\begin{figure}[t]
  \centering
  \includegraphics[width=0.95\columnwidth]{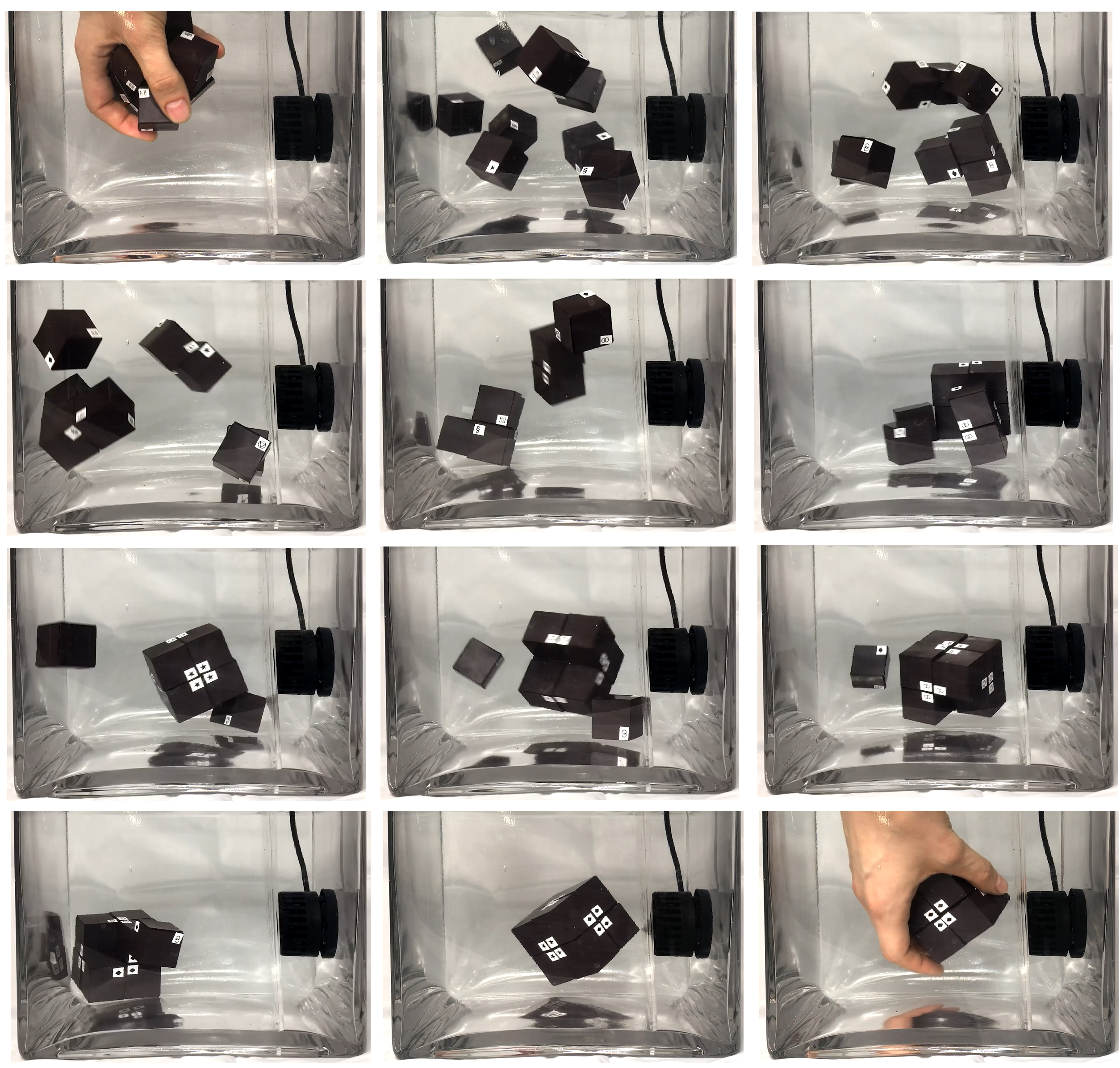}
  \caption{Stochastic self-assembly from (top left) a random arrangement of 8 cubes to a (bottom right) meta cube.}
  \label{fig:assembly}
\end{figure}

We built and programmed 8 cubes to stochastically self-assemble into a "meta cube" (Fig. \ref{fig:assembly}), to suggest how cubes could be programmed to assemble structures into recursively larger cubes. In this configuration, each cube occupies a vertex in an octree and connects to 3 other cubes, yielding an assembly that requires 12 pairs of mating encodings. To do this, we selected one of the 4 maximal cliques (size 12) of mutually compatible encodings (Fig. \ref{fig:teaser} above, right). We wrote a script to translate these encodings into G-code and deployed this on our magnetic plotter, programming each module face in 2 minutes per face. We released all cubes into a glass container (cubic, 200mm side length) filled with tap water, that was stochastically perturbed by a hydraulic pump (Hygger WaveMaker 1600gph). The pump was programmed to produce stochastic flows of random magnitude and frequency to stimulate brownian motion of the cubes. We inserted a laser-cut mesh between the cubes and the pump to promote turbulent flow and to prevent cubes from becoming drawn into the pump inlet. We experimentally calibrated the force of our stochastic disturbance to exceed the attractive force (-0.36) of misassemblies until no permanent misassemblies were observed. Following this procedure, the cubes acquired their correct positions to self-assemble the structure in 32 hours (see supplemental video). After assembly, we re-programmed all faces with new encodings to acquire different final target shapes and measured individual mate forces, observing no difference in the strength of individual mates after reprogramming.

\section{Discussion}



In this paper, we introduced a method to build and program modules capable of selective self-assembly. These modules are scalable; they are passive, inexpensive (\$0.23/module) and consist of COTS bulk material. We have introduced a method of generating encodings that are provably selective. We demonstrate a method of generating highly selective cliques of mutually compatible face encodings for modules, and place guarantees on agnosticism for non-mating configurations in translation and rotation, verifying these results experimentally by instantiating encodings as magnetic polarities. We have shown that these modules can be programmed with encodings that result in specified self-assembled geometries using a simple magnetic plotter, and demonstrated that modules can be re-programmed for new target shapes. Finally, we demonstrated self-assembly using 8 modules in tap water. While our technique was successful at self-assembling conservatively sized systems, a number of limitations and avenues for future work present themselves.

The high selectivity of these encodings\textemdash agnosticism in translation and rotation\textemdash result in significant assembly times, as it constrains the influence of its near field force to areas directly above module faces. This diminishes the ability of programmed faces to easily attract their mates over distances, requiring them to enter each others' narrow volumes of magnetic influence before mutual attraction results in a bond. First, the water chamber size and flow rate could be better calibrated. Future work could investigate adapting the matrices to exhibit larger volumes of influence in the form of magnetic potential wells, with gradients of attraction that help direct mates to their correct orientation. In addition, these modules may be a promising candidate to be used in conjunction with semi-directed stochastic assembly methods \cite{tolley2008dynamically,tolley2011programmable,tolley2010fluidic}, which manipulate the fluid to guide mating modules into their basins of attraction. This may also help address the current system's inability to control assembly order, which would be crucial for complex target geometries. 


Using COTS fridge magnet, the attractive pressure between our mating faces is relatively conservative (250 Pa). Future work could investigate the use of more strongly magnetizable materials to increase mating strength. These encodings could equivalently be used to build strong interfaces by replacing programmed pixels with arrangements of permanent magnets, and for active self-assembly using electromagnets~\cite{nisser2021programmable} or electropermanent magnets whose encodings could be changed online. In addition, the programming step may be rapidly accelerated using parallel arrays of electromagnets to program every pixel simultaneously.   

Our search procedure successfully founded matrices that permitted self-assembly for matrices of order N=8, however future work could improve our search to extend to larger cliques and matrix orders in shorter time. While we have demonstrated the implementation of selectively mating encodings magnetically, future work could also investigate the encoding of these matrices in other binary media, such as in the patterning of electrostatic charge, wettability or chemical bonding, that may permit self-assembly at smaller scales.

\bibliographystyle{IEEEtran}
\bibliography{references}

\begin{thebibliography}{10}
\providecommand{\url}[1]{#1}
\csname url@samestyle\endcsname
\providecommand{\newblock}{\relax}
\providecommand{\bibinfo}[2]{#2}
\providecommand{\BIBentrySTDinterwordspacing}{\spaceskip=0pt\relax}
\providecommand{\BIBentryALTinterwordstretchfactor}{4}
\providecommand{\BIBentryALTinterwordspacing}{\spaceskip=\fontdimen2\font plus
\BIBentryALTinterwordstretchfactor\fontdimen3\font minus
  \fontdimen4\font\relax}
\providecommand{\BIBforeignlanguage}[2]{{%
\expandafter\ifx\csname l@#1\endcsname\relax
\typeout{** WARNING: IEEEtran.bst: No hyphenation pattern has been}%
\typeout{** loaded for the language `#1'. Using the pattern for}%
\typeout{** the default language instead.}%
\else
\language=\csname l@#1\endcsname
\fi
#2}}
\providecommand{\BIBdecl}{\relax}
\BIBdecl

\bibitem{whitesides1991molecular}
G.~M. Whitesides, J.~P. Mathias, and C.~T. Seto, ``Molecular self-assembly and
  nanochemistry: a chemical strategy for the synthesis of nanostructures,''
  \emph{Science}, vol. 254, no. 5036, pp. 1312--1319, 1991.

\bibitem{grzybowski2003electrostatic}
B.~A. Grzybowski, A.~Winkleman, J.~A. Wiles, Y.~Brumer, and G.~M. Whitesides,
  ``Electrostatic self-assembly of macroscopic crystals using contact
  electrification,'' \emph{Nature materials}, vol.~2, no.~4, pp. 241--245,
  2003.

\bibitem{nisser2016feedback}
M.~E. Nisser, S.~M. Felton, M.~T. Tolley, M.~Rubenstein, and R.~J. Wood,
  ``Feedback-controlled self-folding of autonomous robot collectives,'' in
  \emph{2016 IEEE/RSJ International Conference on Intelligent Robots and
  Systems (IROS)}.\hskip 1em plus 0.5em minus 0.4em\relax IEEE, 2016, pp.
  1254--1261.

\bibitem{papadopoulou2017self}
A.~Papadopoulou, J.~Laucks, and S.~Tibbits, ``From self-assembly to
  evolutionary structures,'' \emph{Architectural Design}, vol.~87, no.~4, pp.
  28--37, 2017.

\bibitem{nisser2021laserfactory}
M.~Nisser, C.~C. Liao, Y.~Chai, A.~Adhikari, S.~Hodges, and S.~Mueller,
  ``Laserfactory: A laser cutter-based electromechanical assembly and
  fabrication platform to make functional devices \& robots,'' in
  \emph{Proceedings of the 2021 CHI Conference on Human Factors in Computing
  Systems}, 2021, pp. 1--15.

\bibitem{hauser2020kubits}
S.~Hauser, M.~Mutlu, and A.~J. Ijspeert, ``Kubits: solid-state
  self-reconfiguration with programmable magnets,'' \emph{IEEE Robotics and
  Automation Letters}, vol.~5, no.~4, pp. 6443--6450, 2020.

\bibitem{nisser2017electromagnetically}
M.~Nisser, D.~Izzo, and A.~Borggraefe, ``An electromagnetically actuated,
  self-reconfigurable space structure,'' 2017.

\bibitem{nisser2022electrovoxel}
M.~Nisser, L.~Cheng, Y.~Makaram, R.~Suzuki, and S.~Mueller, ``Electrovoxel:
  Electromagnetically actuated pivoting for scalable modular
  self-reconfigurable robots,'' in \emph{2022 International Conference on
  Robotics and Automation (ICRA)}.\hskip 1em plus 0.5em minus 0.4em\relax IEEE,
  2022, pp. 4254--4260.

\bibitem{romanishin2013m}
J.~W. Romanishin, K.~Gilpin, and D.~Rus, ``M-blocks: Momentum-driven, magnetic
  modular robots,'' in \emph{2013 IEEE/RSJ International Conference on
  Intelligent Robots and Systems}.\hskip 1em plus 0.5em minus 0.4em\relax IEEE,
  2013, pp. 4288--4295.

\bibitem{romanishin20153d}
J.~W. Romanishin, K.~Gilpin, S.~Claici, and D.~Rus, ``3d m-blocks:
  Self-reconfiguring robots capable of locomotion via pivoting in three
  dimensions,'' in \emph{Robotics and Automation (ICRA), 2015 IEEE
  International Conference on}.\hskip 1em plus 0.5em minus 0.4em\relax IEEE,
  2015, pp. 1925--1932.

\bibitem{baca2014modred}
J.~Baca, S.~Hossain, P.~Dasgupta, C.~A. Nelson, and A.~Dutta, ``Modred:
  Hardware design and reconfiguration planning for a high dexterity modular
  self-reconfigurable robot for extra-terrestrial exploration,'' \emph{Robotics
  and Autonomous Systems}, vol.~62, no.~7, pp. 1002--1015, 2014.

\bibitem{zykov2007experiment}
V.~Zykov, H.~Lipson \emph{et~al.}, ``Experiment design for stochastic
  three-dimensional reconfiguration of modular robots,'' in \emph{IEEE Int.
  Conf. Intell. Robots Syst., Self-Reconfigurable Robot. Workshop, San Diego,
  CA}.\hskip 1em plus 0.5em minus 0.4em\relax Citeseer, 2007.

\bibitem{tibbits2012self}
S.~Tibbits, ``The self-assembly line,'' 2012.

\bibitem{hacohen2015meshing}
A.~Hacohen, I.~Hanniel, Y.~Nikulshin, S.~Wolfus, A.~Abu-Horowitz, and
  I.~Bachelet, ``Meshing complex macro-scale objects into self-assembling
  bricks,'' \emph{Scientific reports}, vol.~5, no.~1, pp. 1--8, 2015.

\bibitem{bowden1997self}
N.~Bowden, A.~Terfort, J.~Carbeck, and G.~M. Whitesides, ``Self-assembly of
  mesoscale objects into ordered two-dimensional arrays,'' \emph{Science}, vol.
  276, no. 5310, pp. 233--235, 1997.

\bibitem{lu2021enumeration}
Y.~Lu, A.~Bhattacharjee, D.~Biediger, M.~Kim, and A.~T. Becker, ``Enumeration
  of polyominoes \& polycubes composed of magnetic cubes,'' in \emph{2021
  IEEE/RSJ International Conference on Intelligent Robots and Systems
  (IROS)}.\hskip 1em plus 0.5em minus 0.4em\relax IEEE, 2021, pp. 6977--6982.

\bibitem{jilek2020centimeter}
M.~J{\i}lek, M.~Kulich, and L.~Preucil, ``Centimeter-scaled self-assembly: A
  preliminary study,'' in \emph{Proceedings of the 17th International
  Conference on Informatics in Control, Automation and Robotics (ICINCO).
  ScitePress}, 2020, pp. 438--445.

\bibitem{tolley2008dynamically}
M.~T. Tolley, M.~Krishnan, D.~Erickson, and H.~Lipson, ``Dynamically
  programmable fluidic assembly,'' \emph{Applied Physics Letters}, vol.~93,
  no.~25, p. 254105, 2008.

\bibitem{tolley2010fluidic}
M.~T. Tolley and H.~Lipson, ``Fluidic manipulation for scalable stochastic 3d
  assembly of modular robots,'' in \emph{2010 IEEE international conference on
  robotics and automation}.\hskip 1em plus 0.5em minus 0.4em\relax IEEE, 2010,
  pp. 2473--2478.

\bibitem{tolley2011programmable}
M.~Tolley and H.~Lipson, ``Programmable 3d stochastic fluidic assembly of
  cm-scale modules,'' in \emph{2011 IEEE/RSJ International Conference on
  Intelligent Robots and Systems}.\hskip 1em plus 0.5em minus 0.4em\relax IEEE,
  2011, pp. 4366--4371.

\bibitem{krishnan2008increased}
M.~Krishnan, M.~T. Tolley, H.~Lipson, and D.~Erickson, ``Increased robustness
  for fluidic self-assembly,'' \emph{Physics of Fluids}, vol.~20, no.~7, p.
  073304, 2008.

\bibitem{kalontarov2010hydrodynamically}
M.~Kalontarov, M.~T. Tolley, H.~Lipson, and D.~Erickson, ``Hydrodynamically
  driven docking of blocks for 3d fluidic assembly,'' \emph{Microfluidics and
  Nanofluidics}, vol.~9, no.~2, pp. 551--558, 2010.

\bibitem{jilek2021towards}
M.~J{\'\i}lek, M.~Somr, M.~Kulich, J.~Zeman, and L.~P{\v{r}}eu{\v{c}}il,
  ``Towards a passive self-assembling macroscale multi-robot system,''
  \emph{IEEE Robotics and Automation Letters}, vol.~6, no.~4, pp. 7293--7300,
  2021.

\bibitem{tsutsumi2007multistate}
D.~Tsutsumi and S.~Murata, ``Multistate part for mesoscale self-assembly,'' in
  \emph{SICE Annual Conference 2007}.\hskip 1em plus 0.5em minus 0.4em\relax
  IEEE, 2007, pp. 890--895.

\bibitem{miyashita2009influence}
S.~Miyashita, Z.~Nagy, B.~J. Nelson, and R.~Pfeifer, ``The influence of shape
  on parallel self-assembly,'' \emph{Entropy}, vol.~11, no.~4, pp. 643--666,
  2009.

\bibitem{nisser2022stochastic}
M.~Nisser, Y.~Makaram, and S.~Mueller, ``Stochastic self-assembly with
  magnetically re-programmable voxels,'' 2022.

\bibitem{CorrMag}
``Correlated magnetics,'' \url{http://www.polymagnet.com/}, accessed:
  2021-11-25.

\bibitem{bron1973algorithm}
C.~Bron and J.~Kerbosch, ``Algorithm 457: finding all cliques of an undirected
  graph,'' \emph{Communications of the ACM}, vol.~16, no.~9, pp. 575--577,
  1973.

\bibitem{hagberg2008exploring}
A.~Hagberg, P.~Swart, and D.~S~Chult, ``Exploring network structure, dynamics,
  and function using networkx,'' Los Alamos National Lab.(LANL), Los Alamos, NM
  (United States), Tech. Rep., 2008.

\bibitem{nisser2021programmable}
M.~Nisser, L.~Cheng, Y.~Makaram, R.~Suzuki, and S.~Mueller, ``Programmable
  polarities: Actuating interactive prototypes with programmable
  electromagnets,'' in \emph{The Adjunct Publication of the 34th Annual ACM
  Symposium on User Interface Software and Technology}, 2021, pp. 121--123.

\end{thebibliography}

\end{document}